\documentclass[conference]{IEEEtran}
\IEEEoverridecommandlockouts
\usepackage{cite}
\usepackage{amsmath,amssymb,amsfonts}
\usepackage{algorithmic}
\usepackage{graphicx}
\usepackage{textcomp}
\usepackage{xcolor}
\usepackage{booktabs}
\usepackage{multirow}
\usepackage{pifont}
\usepackage{makecell}
\usepackage{xcolor}

\def\BibTeX{{\rm B\kern-.05em{\sc i\kern-.025em b}\kern-.08em
    T\kern-.1667em\lower.7ex\hbox{E}\kern-.125emX}}
\begin{document}

\title{\huge{GTPC-SSCD: Gate-guided Two-level Perturbation Consistency-based Semi-Supervised Change Detection}}

\author{
    \IEEEauthorblockN{
        Yan Xing$^{1}$\textsuperscript{†}, 
        Qi'ao Xu$^{2}$\textsuperscript{†}, 
        Zongyu Guo$^{3}$,
        Rui Huang$^{3}$\textsuperscript{\textasteriskcentered},
        Yuxiang Zhang$^{3}$
    }
    \IEEEauthorblockA{
        $^{1}$College of Safety Science and Engineering, Civil Aviation University of China, Tianjin, 300300, China \\
        $^{2}$School of Computer Science and Technology, East China Normal University, Shanghai, 20062, China \\
        $^{3}$College of Computer Science and Technology, Civil Aviation University of China, Tianjin, 300300, China \\
        Email: 
        \{yxing, 2022052044, rhuang, yxzhang\}@cauc.edu.cn;
        qaxu@stu.ecnu.edu.cn
    }
    \thanks{$^{\ast}$Corresponding author: Rui Huang.}
    \thanks{\textsuperscript{†}These authors contributed equally.}
    \thanks{This work was supported in part by the Scientific Research Program of Tianjin Municipal Education Commission under Grant 2023KJ232.}
}

\maketitle

\begin{abstract}
Semi-supervised change detection~(SSCD) utilizes partially labeled data and abundant unlabeled data to detect differences between multi-temporal remote sensing images. The mainstream SSCD methods based on consistency regularization have limitations. They perform perturbations mainly at a single level, restricting the utilization of unlabeled data and failing to fully tap its potential. In this paper, we introduce a novel Gate-guided Two-level Perturbation Consistency regularization-based SSCD method~(GTPC-SSCD). It simultaneously maintains strong-to-weak consistency at the image level and perturbation consistency at the feature level, enhancing the utilization efficiency of unlabeled data. Moreover, we develop a hardness analysis-based gating mechanism to assess the training complexity of different samples and determine the necessity of performing feature perturbations for each sample. Through this differential treatment, the network can explore the potential of unlabeled data more efficiently. Extensive experiments conducted on six benchmark CD datasets demonstrate the superiority of our GTPC-SSCD over seven state-of-the-art methods.
\end{abstract}


\section{Introduction}
\label{sec:intro}

Semi-supervised change detection~(SSCD) aims to identify pixel-level changes between two images taken from the same scene at different times, using a limited amount of labeled data and a large amount of unlabeled data~\cite{cheng2023change}.  
It has wide applications in various fields, including natural resource monitoring and utilization, 
disaster monitoring and assessment, urban management and development~\cite{lin2023hyperspectral,Holail2024nsr,10446862}.

\begin{figure}[!t]
\graphicspath{{Fig/}}
\centering
\centerline{\includegraphics[width=1.0\linewidth]{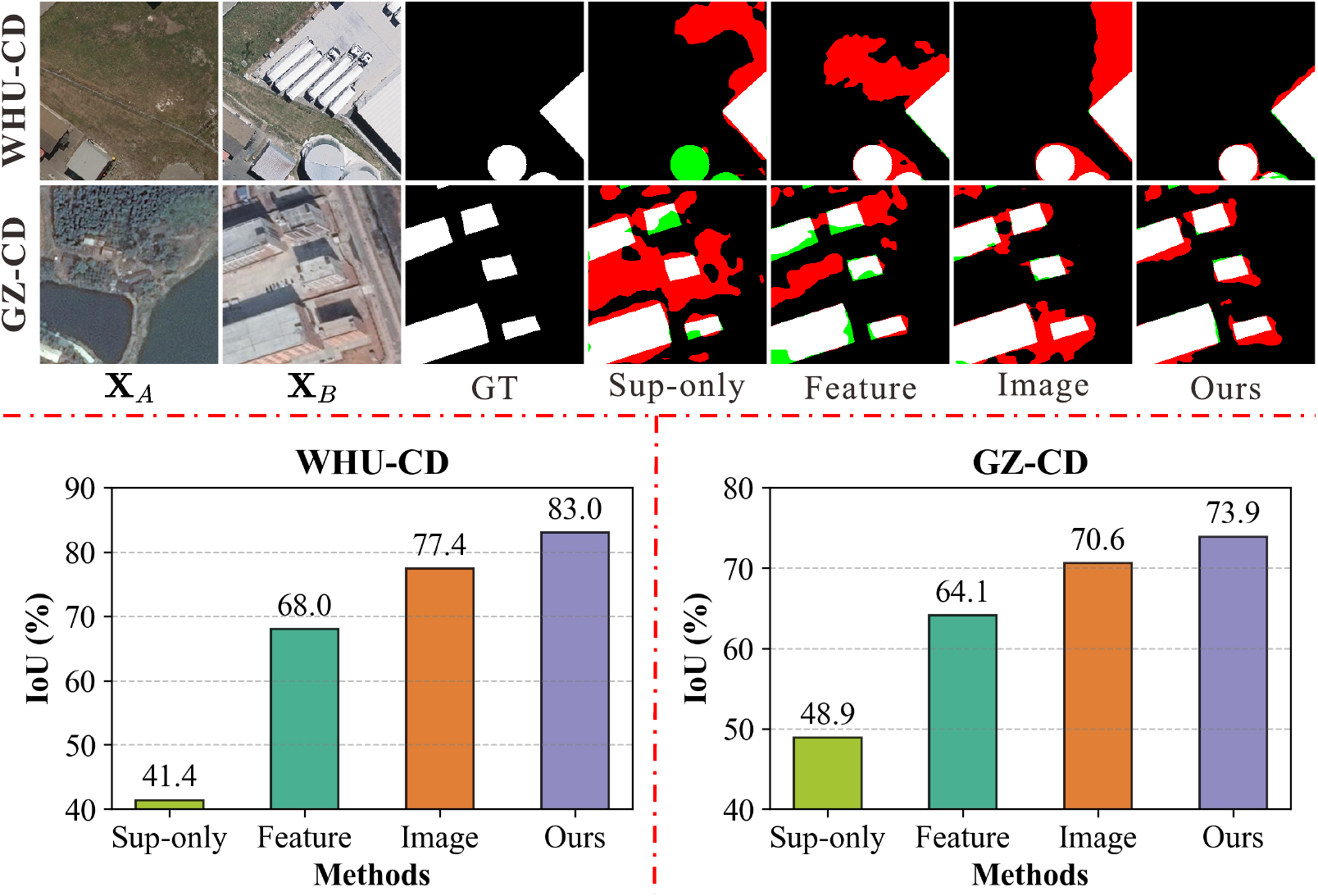}}
\caption{Motivation analysis of SSCD with different perturbation variants under \(5\%\) labeled training data on WHU-CD and GZ-CD datasets. \textbf{Upper part:} Visual cases of different variants. Here, white for TP, black for TN, red for FP, and green for FN. \textbf{Lower part:} Performance comparison of different variants. Sup-only: our method uses only labeled data. Feature: Feature-level perturbation consistency. Image: Image-level strong-to-weak consistency. Ours: Gate-guided two-level perturbation consistency.}
\vspace{-4px}
\label{fig:motivation}
\end{figure}

Existing SSCD methods can be broadly divided into three categories: adversarial learning-based, pseudo-label-based, and consistency regularization-based.
AdvEnt~\cite{vu2019advent}, SemiCDNet~\cite{peng2020semicdnet}, and SALCD~\cite{yang2022change} are typical adversarial learning-based methods that employ alternative optimization strategies to improve the representation learning of their respective models.
Pseudo-label-based methods, such as RC-CD~\cite{wang2022reliable}, SemiSiROC~\cite{kondmann2023semisiroc}, and DCF-CFe~\cite{hou2023deep}, emphasize the improvement of pseudo-label quality and apply some techniques to enhance feature distinctiveness.
Consistency regularization-based methods, including SemiSANet~\cite{sun2022semisanet}, SemiBuildingChange~\cite{sun2023semibuildingchange}, and SemiPTCD~\cite{mao2023semi}, enforce that perturbed images or features produce identical outputs as the original inputs~\cite{sohn2020fixmatch}.

Recently, the consistency regularization framework has gained increasing attention in SSCD due to its simplicity and stability.
However, existing methods mainly focus on modifying network architectures or incorporating supplementary information based on weak-to-strong consistency.
For instance, SemiSANet builds a siamese nested UNet with graph attention, MTCNet~\cite{shu2022mtcnet} augments the change labels with additional segmentation labels, and ST-RCL~\cite{zhang2023strcl} combines self-training and consistency learning.
Meanwhile, feature-level consistency remains relatively simplistic and underdeveloped, such as UniMatch~\cite{yang2023revisiting} applying dropout for feature consistency. 
Thus, there exists substantial potential in integrating image-level and feature-level consistency regularization to enhance SSCD's performance and robustness.   

As our research indicates, simply employing a basic SSCD architecture and applying perturbations only at the image or feature level fails to achieve satisfactory results.
Fig.~\ref{fig:motivation} shows the results of networks employing perturbations at different perspectives, using \(5\%\) labeled data and \(95\%\) unlabeled data on two CD datasets~\cite{ji2018fully,peng2020semicdnet}. The visual cases and performance comparison strongly prove this viewpoint.  
Although perturbations contribute to enhancing model robustness, relying solely on image perturbations may affect pixel integrity and segmentation, and exclusive dependence on feature perturbations might struggle with image diversity.

In this paper, we explore consistency regularization at a broader scope and propose a Gate-guided Two-level Perturbation Consistency regularization-based SSCD method (GTPC-SSCD).
By learning both image-level strong-to-weak consistency and feature-level perturbation consistency, GTPC-SSCD can exploit the unlabeled data more effectively, thereby achieving a more thorough comprehension of the data characteristics. 
Nevertheless, performing the same treatment on all samples constrains the potential efficacy of perturbation, particularly evident in remote sensing change detection scenarios.
To alleviate this issue, we develop a hardness analysis-based gating mechanism. It is designed to distinguish the difficulty of training samples and determine the application of feature perturbation accordingly.
We carry out comparative experiments on six CD datasets  with seven state-of-the-art methods, and the proposed GTPC-SSCD achieves the best performance. 

Our contributions are summarized as follows:
\begin{itemize}
    \item We separately analyze image-level and feature-level perturbation implementations in SSCD  and combine them with adaptive strategies to improve performance.
    \item We propose GTPC-SSCD, a Gate-guided Two-level Perturbation Consistency regularization-based SSCD method. And we design a hardness analysis-based gating mechanism to assess the difficulty of samples and determine the necessity of applying feature perturbations.
    \item We conduct extensive experiments on six benchmark CD datasets and our method exhibits higher accuracy than other seven state-of-the-art methods.
\end{itemize}



\section{Methodology}

Semi-supervised change detection~(SSCD) utilizes a limited amount of labeled data and a large amount of unlabeled data to train a change detection network for accurate change maps generation. 
The training set consists of two subsets, a labeled set and an unlabeled set. 
The labeled set can be represented as
$\mathcal{D}_l= \{{(\mathbf{X}^l_{A,i},\mathbf{X}^l_{\mathrm{B},i}), \mathbf{Y}^l_i}\}^M_{i=1}$,
where $(\mathbf{X}^l_{A,i},\mathbf{X}^l_{B,i})$ denotes the $i$-th labeled image pair, $\mathbf{X}^l_{A,i}$ is the pre-change image, $\mathbf{X}^l_{B,i}$ is the post-change image, and $\mathbf{Y}^l_i$ is the corresponding ground truth. 
Let $\mathcal{D}_u=\{{(\mathbf{X}^u_{A,i}, \mathbf{X}^u_{B,i})}\}^N_{i=1}$ denotes the unlabeled set. 
$(\mathbf{X}^u_{A,i},\mathbf{X}^u_{B,i})$ is the $i$-th unlabeled image pair. 
$M$ and $N$ indicate the number of labeled and unlabeled image pairs, respectively. 
In most cases, we have $N>>M$.
In the following sections, we will detail the proposed GTPC-SSCD.

\subsection{Our GTPC-SSCD method} 
\label{ssec:consistency_training}

Our method consists of a supervised training part and an unsupervised training part.

\begin{figure}[!htb]
\graphicspath{{Fig/}}
\centering
\centerline{\includegraphics[width=1.0\linewidth]{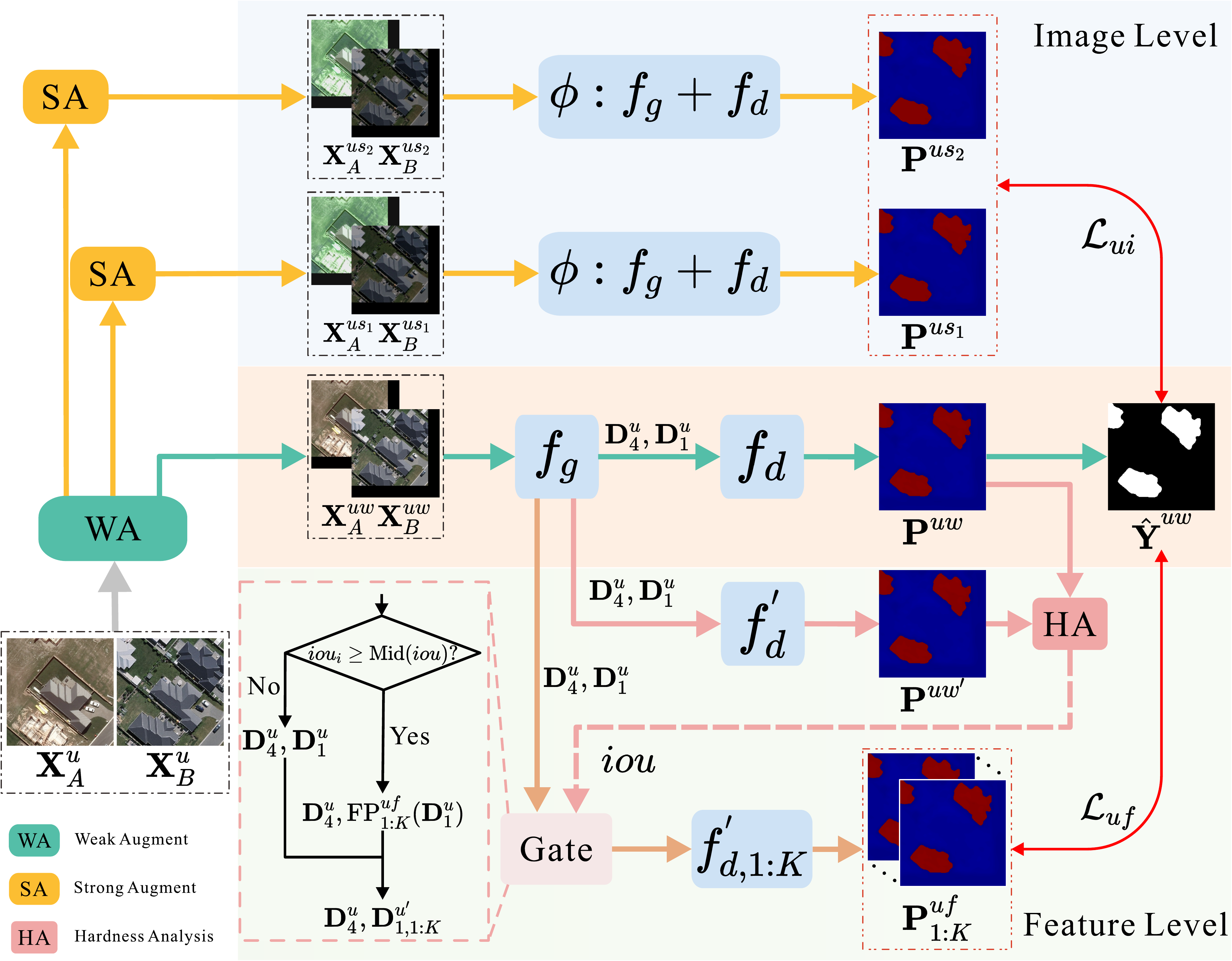}}
\caption{Framework of the proposed GTPC-SSCD method. }
\vspace{-4px}
\label{fig:framework}
\end{figure}

\subsubsection{Supervised training part} 
We use the labeled set $\mathcal{D}_l$ to train the CD network $\phi$. The network processes weakly augmented image pairs to generate change map $\mathbf{P}^l$.
We utilize the standard cross-entropy (CE) loss as supervision. The supervised loss $\mathcal{L}_s$ is formulated as:
\begin{align}
    \mathcal{L}_s= \mathcal{L}_{CE}(\mathbf{P}^{l},\mathbf{Y}^l),
\end{align}
where $\mathcal{L}_{CE}$ represents the CE loss, which measures the dissimilarity between the predicted change map $\mathbf{P}^l$ and the ground truth $\mathbf{Y}^l$.

\subsubsection{Unsupervised training part} 
As illustrated in Fig.~\ref{fig:framework}, we apply two-level perturbations to train $\phi$ using the unlabeled set $\mathcal{D}_u$.
Specifically, we enforce strong-to-weak consistency at the image level and maintain perturbation consistency at the feature level. 
The former enhances the model's robustness against various data transformations like illumination and season changes, while the latter sharpens the model's generalization and interference resistance.  %

\textbf{Image-level strong-to-weak consistency learning.} 
Drawing inspiration from ConMatch~\cite{kim2022conmatch}, we expand the quantity of strong augmentation branches, and enforce the strong-to-weak consistency by aligning strong and weak outputs. 
It promotes a more efficient exploitation of the latent potential within unlabeled data, strengthening the model's adaptability across diverse data scenarios and enhancing its robustness.
Specifically, we apply two independent strong augmentations on $(\mathbf{X}^{uw}_A,\mathbf{X}^{uw}_B)$, resulting in $(\mathbf{X}^{us_1}_{A},\mathbf{X}^{us_1}_{B})$ and $(\mathbf{X}^{us_2}_{A},\mathbf{X}^{us_2}_{B})$.
They are fed into $\phi$ and generate three change maps $\mathbf{P}^{uw}$, $\mathbf{P}^{us_1}$ and $\mathbf{P}^{us_2}$.

We use $\mathbf{P}^{uw}$ to generate pseudo-label $\hat{\mathbf{Y}}^{uw}$ by:
\begin{equation}
    \hat{\mathbf{Y}}^{uw}=  
    \begin{cases}
 1, &  if \ \mathbf{P}^{uw} > \tau  \\
 0, &  else  
\end{cases}
\end{equation} 
where $\tau$ = 0.95 is a confidence threshold.
The image-level consistency loss $\mathcal{L}_{ui}$ is expressed as:
\begin{equation}
    \mathcal{L}_{ui} = \frac{1}{2} ( \mathcal{L}_{CE}(\mathbf{P}^{us_1},\hat{\mathbf{Y}}^{uw}) + \mathcal{L}_{CE}(\mathbf{P}^{us_2},\hat{\mathbf{Y}}^{uw}) ) .
\end{equation} 
Here, weak augmentations consist of random resizing, cropping, and horizontal flipping. 
Strong augmentations include random color jittering, Gaussian blur, and CutMix. 

\textbf{Feature-level perturbation consistency learning.}
To establish feature-level consistency, we implement several different feature perturbations on change features and ensure the perturbed outputs align with the originals. 
%
%
The process of generating change maps $\mathbf{P}^{uf}_{1:K}$ is formulated as:
\begin{align}
\mathbf{P}^{uf}_{1:K}= f^{'}_{d,1:K}( \mathbf{D}^u_4, \mathbf{D}^{u'}_{1,1:K} ),
\end{align} 
where $\mathbf{D}^{u'}_{1,1:K} = \{\mathbf{D}^{u'}_{1,k}\}^K_{k=1}$, represents the difference features after perturbation. $f^{'}_{d,1:K}= \{f^{'}_{d,k}\}^K_{k=1}$, denote the auxiliary decoders. $\mathbf{P}^{uf}_{1:K}=\{\mathbf{P}^{uf}_k\}^K_{k=1}$.
The feature-level consistency loss $\mathcal{L}_{uf}$ is expressed as:
\begin{equation}
\mathcal{L}_{uf} = 
\frac{1}{K} \sum^{K}_{k=1} \mathcal{L}_{CE}(\mathbf{P}^{uf}_k,\hat{\mathbf{Y}}^{uw}) ,
\end{equation} 
where $K$ is the total number of the auxiliary decoders.
We adopt seven types of feature perturbations~\cite{ouali2020semi}, i.e., feature noise, feature dropout, object masking, context masking, guided cutout, Intermediate VAT, and random dropout.

\textbf{Hardness analysis-based gating mechanism.}
To boost the model's adaptability to samples with varying complexity, we develop a hardness-based gating mechanism.
It utilizes an additional decoder to generate a change map $\mathbf{P}^{uw'}_i$. Then, an Intersection over Union (IoU) comparison is performed between $\mathbf{P}^{uw'}_i$ and $\mathbf{P}^{uw}_i$ to quantitatively assess the complexity of each sample.
Utilizing the gating mechanism, we can select challenging samples and apply various perturbations on $\mathbf{D}^u_1$ of these selected samples to obtain perturbed features. The 
whole process can be formulated as:
\begin{equation}
    iou_i = \mathrm{IoU} (\mathbf{P}^{uw'}_i,\mathbf{P}^{uw}_i ), 
\end{equation}
\begin{equation}
    \mathbf{D}^{u'}_{1,1:K} =  
    \begin{cases}
    \mathrm{FP}_{1:K}(\mathbf{D}^u_1) , & if \ iou_i \ge \mathrm{Mid}(iou)  \\
    \mathbf{D}^u_1, &  else   
    \end{cases}
\end{equation} 
where $iou_i$ denotes the IoU score of the $i$-th sample.
$\mathrm{Mid}(iou)$ is calculated as the median value, serving as a stable and reliable ratio. $\mathrm{FP}_{1:K}(\cdot)$ denotes $K$ different perturbations to the input features.
$\mathbf{D}^{u'}_{1,1:K} = \{\mathbf{D}^{u'}_{1,k}\}^K_{k=1}$, is the set of $\mathbf{D}^{u'}_{1,k}$.
Through this differential sample processing strategy, the model can effectively harness the potential of unlabeled data, enhancing its overall performance.

\subsubsection{Loss function} 
The total loss consists of the supervised loss $\mathcal{L}_{s}$, the image-level consistency loss $\mathcal{L}_{ui}$ and the feature-level consistency loss $\mathcal{L}_{uf}$. It is formulated as:
\begin{equation}
    \mathcal{L} = \lambda_1 \mathcal{L}_{s} + \lambda_2 \mathcal{L}_{ui} + \lambda_3 \mathcal{L}_{uf},
\end{equation}
where $\lambda_1=0.5$, $\lambda_2=0.25$, and $\lambda_3=0.25$. 

\subsection{Implementation detail}

\subsubsection{Change detection network} 
\label{ssec:CDnetwork}
Our CD network $\phi$ consists of two components: a difference feature generator $f_g$ and a decoder $f_d$. The generation of change map $\mathbf{P}$ is expressed by:
\begin{equation}
    \label{eq:network}
    \mathbf{P}= \phi(\mathbf{X}_A,\mathbf{X}_B)= f_d(f_g (\mathbf{X}_{A}, \mathbf{X}_{B})).
\end{equation}

\textbf{Difference feature generator.} 
The feature encoder is built on ResNet50~\cite{he2016deep} with a siamese setup. 
Deep features contain rich semantic information, while shallow features encompass abundant details.
We use the features of the first and fourth residual modules to calculate the difference features $\mathbf{D}_i$ by: 
\begin{equation}
    \mathbf{D}_i= |\mathbf{C}^A_i- \mathbf{C}^B_i |, i=1,4, 
\end{equation}
where $\mathbf{C}^A_i$ and $\mathbf{C}^B_i$ are the features of the $i$-th residual module from $\mathbf{X}_A$ and $\mathbf{X}_B$, respectively.
$|\cdot|$ is the absolute operation.

\textbf{Decoder.}
The decoder is used to process change features and generate change maps. We apply an Atrous Spatial Pyramid Pooling (ASPP)~\cite{chen2018encoder} on $\mathbf{D}_4$ to obtain richer information $\mathbf{F}_4$.
Next, we combine $\mathbf{D}_1$ and the upsampled feature of $\mathbf{F}_4$ and employ a classifier to generate $\mathbf{P}$. The process can be formulated as:
\begin{align}
    \mathbf{F}_4 &= \mathrm{ASPP}( \mathbf{D}_4), \\
    \mathbf{F}_1= \mathrm{CBR}_3( \mathrm{CBR}_3 &( [ \mathrm{Up(\mathbf{F}_4)}, \mathrm{CBR}_1(\mathbf{D}_1) ] )), \\
    \mathbf{P} &= \mathrm{Conv}_1(\mathbf{F}_1),
\end{align}
where $\mathrm{ASPP(\cdot)}$ denotes the ASPP process, $\mathrm{Up}(\cdot)$ is upsampling operation, $[\cdot,\cdot]$ denotes concatenate operation.
$\mathrm{CBR}_k(\cdot)$ denotes a \(k \times k\) convolutional layer followed with Batch Normalization and ReLU, and $\mathrm{Conv}_1(\cdot)$ represents a \(1 \times 1\) convolutional layer.

%

\subsubsection{Super-parameters}
The experiments are implemented using PyTorch on an NVIDIA RTX2080Ti GPU.
Our model adopts the SGD optimizer with a learning rate of 0.02, a momentum of 0.9, and a weight decay of 1e-4. 
All models are trained for 80 epochs with a batch size of 4 for both labeled and unlabeled data.

\section{Experiment}
\label{sec:experiment}

\subsection{Experimental Setup}
\label{ssec:setting}

\textbf{Baselines.}
To validate the effectiveness and superiority of our proposed method, we compare it with seven existing state-of-the-art SSCD methods, including AdvEnt~\cite{vu2019advent}, s4GAN~\cite{mittal2019semi}, SemiCDNet~\cite{peng2020semicdnet}, SemiCD~\cite{bandara2022revisiting}, RC-CD~\cite{wang2022reliable}, SemiPTCD~\cite{mao2023semi}, and UniMatch~\cite{yang2023revisiting}.
Notably, the first three are adversarial learning-based methods, RC-CD is a pseudo-label-based method, and the remaining three are based on consistency regularization.
Sup-only refers to our method trained only on labeled data.
All methods are conducted with PyTorch and trained on the same training sets.

\begin{table*}[!htbp]
\centering
\caption{Dataset details.}
\vspace{-2px}
\label{tab:dataset}
\resizebox{0.78\textwidth}{!}{
\begin{tabular}{c | c c c c c}
\toprule
    Dataset & Image pairs & Image size & Train/Val/Test & Spatial resolution &  Changes \\
\midrule
    WHU-CD & 1 & $15354\times 32507\times 3$ & $5974/743/744$ & 0.075 m/pixel & building   \\
    LEVIR-CD & 637 & $1024\times 1024\times 3$ & $7120/1024/2048$ & 0.5 m/pixel & \makecell{building, villa residence, \\ apartment, garage, \textit{etc}.} \\
    BCD & 1922 & $256\times 256\times 3$ & $1538/192/192$ & 0.2 m/pixel & building   \\
    GZ-CD & 19 & \makecell[c]{\(1006\times 1168 \times 3,\) \\ \(4936\times 5224 \times 3\) } & $2882/360/361$ & 0.55 m/pixel & \makecell[c]{building, road, forests, \\ farmlands, \textit{etc}.}  \\
    EGY-BCD & 6091 & $256\times 256\times 3$ & $4264/1218/609$ & 0.25 m/pixel & building  \\
    CDD & 16000 & $256\times 256\times 3$ & $10000/3000/3000$ & 0.03-1 m/pixel & \makecell[c]{building, car, \\ tree, road, \textit{etc}. } \\
\bottomrule
\end{tabular}
}
\vspace{-2px}
\end{table*}

\begin{table*}[!htb]
\centering
\caption{Quantitative comparison of different methods on six CD datasets. The highest scores are in \textbf{bold}, and the second are \underline{underlined}.}
\vspace{-2px}
\resizebox{0.95\linewidth}{!}{
\label{table:main_result1}
\begin{tabular}{c|cc|cc|cc|cc|cc|cc|cc|cc}
\toprule
\multirow{3}{*}{Method} & \multicolumn{8}{c|}{\textbf{WHU-CD}} & \multicolumn{8}{c}{\textbf{LEVIR-CD}} \\ 
  \multirow{2}{*}{} & \multicolumn{2}{c|}{5\%} & \multicolumn{2}{c|}{10\%} & \multicolumn{2}{c|}{20\%} & \multicolumn{2}{c|}{40\%} & \multicolumn{2}{c|}{5\%} & \multicolumn{2}{c|}{10\%} & \multicolumn{2}{c|}{20\%} & \multicolumn{2}{c}{40\%} \\ 
 & IoU & OA & IoU & OA & IoU & OA & IoU & OA & IoU & OA & IoU & OA & IoU & OA & IoU & OA \\
\midrule
AdvEnt~\cite{vu2019advent} & 57.7 &	97.87 &	60.5 & 97.79 & 69.5 & 98.50 & 76.0 & 98.91 
& 67.1 & 98.15 & 70.8 & 98.38 & 74.3 & 98.59 & 75.9 & 98.67  \\ 
s4GAN~\cite{mittal2019semi} & 57.3 & 97.94 & 58.0 & 97.81 & 67.0 & 98.41 & 74.3 & 98.85 
& 66.6 & 98.16 & 72.2 & 98.48 & 75.1 & 98.63 & 76.2 & 98.68  \\
SemiCDNet~\cite{peng2020semicdnet} & 56.2 & 97.78 & 60.3 & 98.02 & 69.1 & 98.47 & 70.5 & 98.59 
& 67.4 & 98.11 & 71.5 & 98.42 & 74.9 & 98.58 & 75.5 & 98.63  \\
SemiCD~\cite{bandara2022revisiting} & 65.8 & 98.37 & 68.0 & 98.45 & 74.6 & 98.83 & 78.0 & 99.01 
& 74.2 & 98.59 & 77.1 & 98.74 & 77.9 & 98.79 & 79.0 & 98.84  \\
RC-CD~\cite{wang2022reliable} & 58.0 & 98.01 & 61.7 & 98.00 & 74.0 & 98.83 & 73.9 & 98.85
& 74.0 & 98.52 & 76.1 & 98.65 & 77.1 & 98.70 & 77.6 & 98.72  \\
SemiPTCD~\cite{mao2023semi} & 74.1 & 98.85 & 74.2 & 98.86 & 76.9 & 98.95 & 80.8 & 99.17 & 71.2 & 98.39 & 75.9 & 98.65 & 76.6 & 98.65 & 77.2 & 98.74  \\
UniMatch~\cite{yang2023revisiting} & \underline{78.7} & \underline{99.11} & \underline{79.6} & \underline{99.11} & \underline{81.2} & \underline{99.18} & \underline{83.7} & \underline{99.29} & \underline{82.1} & \underline{99.03} & \underline{82.8} & \underline{99.07} & \underline{82.9} & \underline{99.07} & \underline{83.0} & \underline{99.08}  \\

\midrule
Sup-only & 41.4 & 96.36 & 55.0 & 97.28 & 49.3 & 96.98 & 65.6 & 98.26 & 71.8 & 98.46 & 78.0 & 98.78 & 77.7 & 98.76 & 78.9 & 98.83  \\
Ours & \textbf{83.0} & \textbf{99.30} & \textbf{82.8} & \textbf{99.28} & \textbf{84.2} & \textbf{99.31} & \textbf{85.4} & \textbf{99.37} & \textbf{83.2} & \textbf{99.09} & \textbf{83.1} & \textbf{99.08} & \textbf{83.7} & \textbf{99.12} & \textbf{83.5} & \textbf{99.10}  \\


\midrule

\multirow{1}{*}{} & \multicolumn{8}{c|}{\textbf{BCD}} & \multicolumn{8}{c}{\textbf{GZ-CD}} \\ 
\midrule
AdvEnt~\cite{vu2019advent} & 62.4 & 91.93 & 69.1 & 93.29 & 75.2 & 94.85 & 77.1 & 95.14
& 56.7 & 95.52 & 57.5 & 95.99 & 70.3 & 97.28 & 70.8 & 97.29  \\ 
s4GAN~\cite{mittal2019semi} & 57.7 & 90.77 & 70.9 & 93.95 & 74.9 & 94.64 & 78.0 & 95.35 
& 59.4 & 96.13 & 61.6 & 96.23 & 68.5 & 97.10 & 69.4 & 97.08  \\
SemiCDNet~\cite{peng2020semicdnet} & 59.4 & 91.49 & 70.3 & 93.83 & 74.4 & 94.65 & 77.7 & 95.33 
& 57.9 & 95.38 & 54.9 & 95.52 & 68.9 & 97.16 & 69.7 & 97.20  \\
SemiCD~\cite{bandara2022revisiting} & 60.9 & 92.08 & 73.1 & 94.23 & 77.5 & 95.26 & 79.3 & 95.66 
& 59.5 & 96.27 & 58.6 & 96.03 & 67.0 & 97.03 & 71.5 & 97.36  \\
RC-CD~\cite{wang2022reliable} & 75.4 & 94.79 & 77.7 & 95.18 & 79.7 & 95.77 & 80.6 & 95.95
& 62.2 & 96.26 & 63.9 & 96.55 & \underline{74.1} & 97.69 & \underline{74.2} & \underline{97.57}  \\
UniMatch~\cite{yang2023revisiting} & \underline{78.3} & \underline{96.17} & \underline{79.9} & \underline{96.48} & \underline{80.7} & \underline{96.58} & \underline{81.7} & \underline{96.79} & \underline{68.7} & \underline{97.06} & \underline{69.5} & \textbf{97.41} & 72.8 & \underline{97.71} & 71.1 & 97.48 \\

\midrule
Sup-only & 59.3 & 91.49 & 63.3 & 92.48 & 74.1 & 95.16 & 70.1 & 95.16  
& 48.9 & 94.49 & 36.9 & 91.94 & 53.5 & 95.44 & 54.4 & 95.73 \\
Ours & \textbf{79.9} & \textbf{96.52} & \textbf{81.3} & \textbf{96.76} & \textbf{81.3} & \textbf{96.78} & \textbf{82.4} & \textbf{96.95}  
& \textbf{73.9} & \textbf{97.71} & \textbf{70.7} & \underline{97.36} & \textbf{77.8} & \textbf{98.09} & \textbf{78.4} & \textbf{98.17}    \\

\midrule

\multirow{1}{*}{} & \multicolumn{8}{c|}{\textbf{EGY-BCD}} & \multicolumn{8}{c}{\textbf{CDD}} \\ 

\midrule
AdvEnt~\cite{vu2019advent} 
& 52.0 & 95.30 & 58.1 & 96.26 & 59.8 & 96.46 & 63.8 & 96.94 
& 63.8 & 94.98 & 72.7 & 96.32 & 79.0 & 97.22 & 82.8 & 97.72 \\
s4GAN~\cite{mittal2019semi} 
& 53.2 & 95.62 & 56.5 & 96.26 & 59.4 & 96.62 & 64.1 & 96.83 
& 62.5 & 94.92 & 70.4 & 96.12 & 78.7 & 97.20 & 82.8 & 97.72 \\
SemiCDNet~\cite{peng2020semicdnet} 
& 52.7 & 95.36 & 56.9 & 96.02 & 59.8 & 96.53 & 63.6 & 96.96 
& 64.3 & 95.01 & 72.5 & 95.88 & 79.1 & 97.23 & 82.6 & 97.73 \\
SemiCD~\cite{bandara2022revisiting} 
& 54.3 & 95.79 & 59.2 & 96.29 & 61.8 & 96.61 & 65.4 & 96.96 
& 66.4 & 95.39 & 74.9 & 96.71 & 81.2 & 97.54 & 84.4 & 97.93 \\
RC-CD~\cite{wang2022reliable} 
& 59.0 & 96.17 & 61.6 & 96.51 & \underline{64.6} & 96.79 & \underline{67.7} & 97.09 
& 69.2 & 95.82 & 73.4 & 96.29 & 80.1 & 97.32 & 82.7 & 97.67 \\
UniMatch~\cite{yang2023revisiting} 
& \underline{62.8} & \underline{96.74} & \underline{65.5} & \underline{97.10} & 63.6 & \underline{96.91} & 67.3 & \underline{97.26}
& \underline{74.8} & \underline{96.69} & \underline{80.3} & \underline{97.41} & \underline{86.8} & \underline{98.30} & \textbf{90.7} & \textbf{98.81} \\

\midrule
Sup-only
& 44.5 & 94.03 & 54.9 & 95.65 & 44.8 & 94.28 & 60.0 & 96.37 
& 49.7 & 92.58 & 62.3 & 94.71 & 66.2 & 95.32 & 78.5 & 97.15 \\
Ours 
& \textbf{64.5} & \textbf{96.91} & \textbf{65.9} & \textbf{97.15} & \textbf{66.1} & \textbf{97.13} & \textbf{68.9} & \textbf{97.44} 
& \textbf{76.4} & \textbf{96.89} & \textbf{82.3} & \textbf{97.70} & \textbf{88.0} & \textbf{98.46} & \underline{90.2} & \underline{98.74} \\

\bottomrule

\end{tabular}}
\vspace{-4px}
\end{table*}

\textbf{Datasets.} We conduct experiments on six widely used remote sensing CD datasets, namely WHU-CD~\cite{ji2018fully}, LEVIR-CD~\cite{chen2020spatial}, BCD~\cite{ji2018fully}, GZ-CD~\cite{peng2020semicdnet}, EGY-BCD~\cite{holail2023afde}, and CDD~\cite{lebedev2018change}.
Table~\ref{tab:dataset} shows the details of these datasets, including the number of image pairs, image size, train/val/test splits, spatial resolution, and change types.
GZ-CD and CDD datasets encompass a diverse range of change types, such as buildings, vehicles, and roads, and others. The remaining four CD datasets mainly concentrate on various forms of building alterations.
All images are cropped into non-overlapping patches of size \(256\times 256\), which are then divided into training, validation, and test sets. 
The training set is further divided into labeled and unlabeled data with the following ratios:
\([5\%,95\%]\), \([10\%,90\%]\), \([20\%,80\%]\), \([40\%,60\%]\).

\textbf{Criterion.} 
Following Bandara \textit{et al.}~\cite{bandara2022revisiting} and Yang \textit{et al.}~\cite{yang2023revisiting}, we adopt intersection over union (IoU) and overall accuracy (OA) as evaluation metrics. 
IoU measures the overlap between predicted and actual changes for precision, and OA gives an overall accuracy measure. 
Their formulas are: 
\begin{equation}
\mathrm{IoU} = \mathrm{TP} / (\mathrm{TP + FP + FN})
\end{equation}
\begin{equation}
\mathrm{OA} = (\mathrm{TP + TN}) / (\mathrm{TP + FP + FN + TN})
\end{equation}
where TP, TN, FP, and FN represent true positive, true negative, false positive, and false negative, respectively.

\begin{figure*}[!ht]
\graphicspath{{Fig/}}
\centering
\centerline{\includegraphics[width=0.80\linewidth]{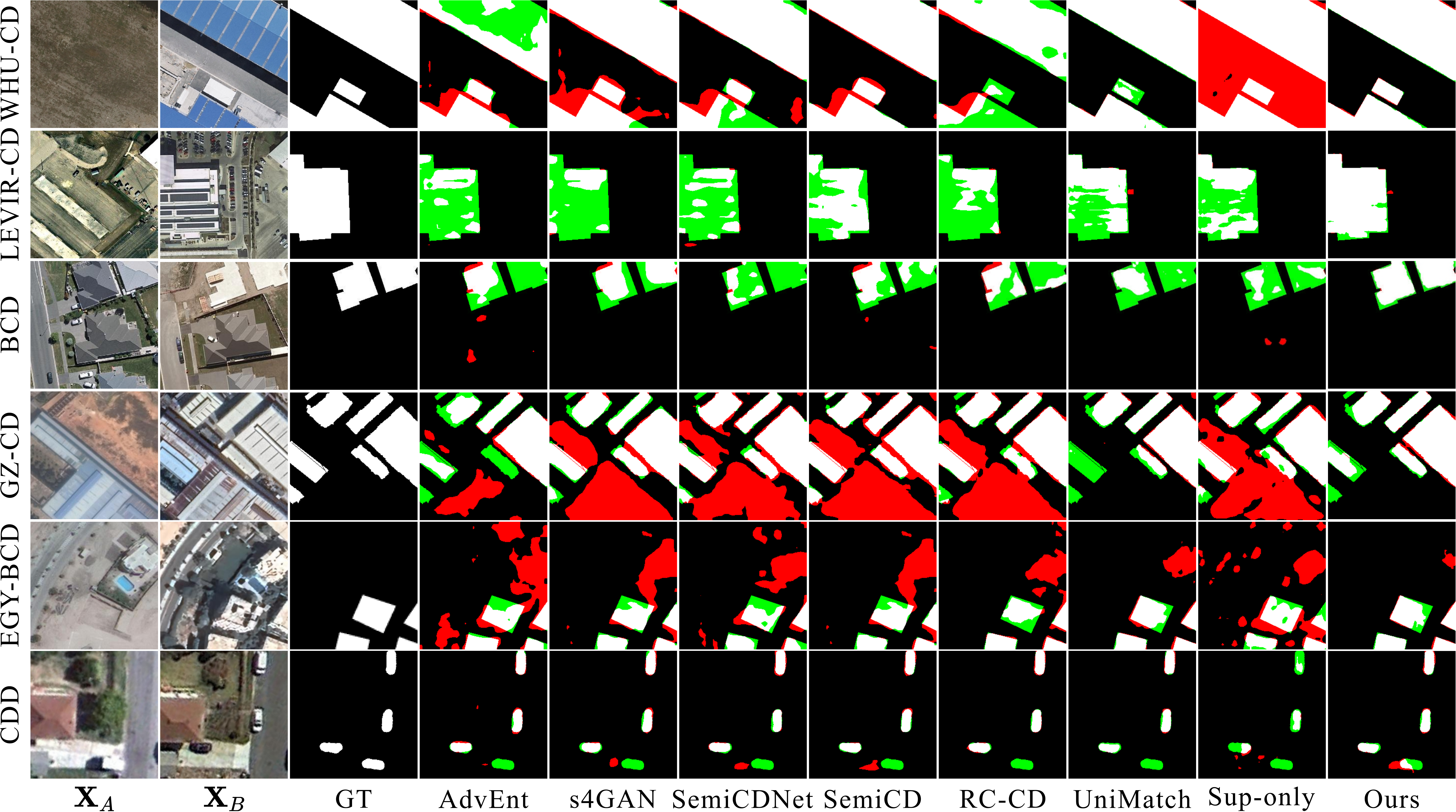}}
\caption{Detection results of different methods on six CD datasets at \(5\%\) labeled ratio. Here, white for TP, black for TN, red for FP, and green for FN.}
\label{fig:visiual_result}
\vspace{-4px}
\end{figure*}

\subsection{Results and Discussion}
\label{ssec:results}
\textbf{Comparison with the state-of-the-art.}
Table~\ref{table:main_result1} exhibits the quantitative results of different methods on WHU-CD, LEVIR-CD, BCD, GZ-CD, EGY-BCD, and CDD datasets.
From the findings in these tables, we can draw the following observations.
\ding{182} Most SSCD methods outperform Sup-only under the same partition, confirming the benefit of unlabeled data for SSCD.
\ding{183} Our method achieves the best performance across all four partitions, notably surpassing other methods, especially on WHU-CD and GZ-CD. 
On WHU-CD, compared to the leading method UniMatch, our method brings \(4.3\%\), \(3.2\%\), \(3.0\%\), and \(1.7\%\) performance gain in terms of IoU with \(5\%\), \(10\%\), \(20\%\), and \(40\%\) labeled training data, respectively.
On GZ-CD, the improved performance with IoU of our method over the best UniMatch are \(5.2\%\), \(1.2\%\), \(5.0\%\), and \(7.3\%\).

\textbf{Detection results.} 
Fig.~\ref{fig:visiual_result} shows some examples of different methods on six CD datasets under \(5\%\) labeled partition. 
Our method exhibits higher accuracy and richer details.
Both quantitative and qualitative results support our method's superiority.

\textbf{Computational complexity analysis.} 
Table~\ref{tab:comparison} details the parameters and computation costs of different methods.
Our method shows 65.87 GFLOPs, 57.3 M parameters, and an FPS of 51.64.
Compared with UniMatch, our method significantly boosts performance with moderate GFLOPs and parameters, balancing performance and efficiency effectively.

\vspace{-4px}

\begin{table}[!htb]
\centering
\caption{Comparison of parameters and computational costs.}
\vspace{-2px}
\label{tab:comparison}
\resizebox{0.28\textwidth}{!}{
\large
\begin{tabular}{c|c c c}
\toprule
    Method & GFLOPs & Params (M) & FPS \\
\midrule
    AdvEnt & 74.06 & 49.6 & 40.82  \\
    s4GAN & 76.72 & 49.6 & 41.54  \\
    SemiCDNet & 75.69 & 52.4 & 40.87 \\
    SemiCD & 75.37 & 50.7 & 39.30  \\
    UniMatch & 38.30 & 40.5 & 52.19  \\
\midrule
    Ours & 65.87 & 57.3 & 51.64  \\
\bottomrule
\end{tabular}}
\vspace{-8px}
\end{table}

%
\begin{table}[!htb]
\centering
\caption{Ablation study on WHU-CD dataset.}
\vspace{-2px}
\label{table:ablation1}
\resizebox{1.0\linewidth}{!}{
\begin{tabular}{c|c|cc|cc|cc|cc}
\toprule
     & \multirow{2}{*}{Method} & \multicolumn{2}{c|}{5\%} & \multicolumn{2}{c|}{10\%} & \multicolumn{2}{c|}{20\%} & \multicolumn{2}{c}{40\%}  \\ 
    & & IoU & OA & IoU & OA & IoU & OA & IoU & OA \\
\midrule
 \multirow{4}{*}{\rotatebox{90}{\shortstack{Without \\ Gate}}}  & 
    Sup-only & 41.4 & 96.36 & 55.0 & 97.28 & 49.3 & 96.98 & 65.6 & 98.26  \\
    & Feature 
    & 68.0 & 98.49 & 73.9 & 98.80 & 74.3 & 98.77 & 82.0 & 99.21  \\
    & Image
    & 77.4 & 99.03 & 80.0 & 99.14 & 77.5 & 98.97 & 82.2 & 99.22  \\
    & Feature + Image & 81.7 & 99.24 & 81.6 & 99.21 & 79.2 & 99.07 & 84.8 & 99.35 \\
\midrule
\multirow{4}{*}{\rotatebox{90}{\shortstack{With \\ Gate}}} 
    & FP(d4) & 79.4 & 99.13 & 78.0 & 99.03 & \underline{82.9} & \underline{99.25} & 83.4 & 99.28  \\
    & FP(d1, d4) & 78.5 & 99.08 & 81.3 & 99.20 & 79.4 & 99.07 & 84.4 & 99.33 \\
    & Ours(FP(d1)) & \textbf{83.0} & \textbf{99.30} & \textbf{82.8} & \textbf{99.28} & \textbf{84.2} & \textbf{99.31} & \underline{85.4} & \underline{99.37}   \\
    & ResNet101 & \underline{82.6} & \underline{99.28} & \underline{82.3} & \underline{99.24} & 82.1 & 99.21 & \textbf{85.7} & \textbf{99.39}   \\
\bottomrule
\end{tabular}}
\end{table}

\textbf{Ablation study.}
Table~\ref{table:ablation1} illustrates four sets of ablation studies. From this table, we have the following findings.
\ding{182} Under the ``Without Gate" condition, both image-level and feature-level perturbations contribute to performance enhancement, and their combined utilization further improves performance. 
\ding{183} A comparison between ``Feature + Image" and ``Ours(FP(d1))" reveals that incorporating the gate mechanism helps distinguish challenging samples, reducing training difficulty and improving performance.
\ding{184} Under the ``With Gate" condition, contrasting ``FP(d4)", ``FP(d1,d4)", and ``Ours(FP(d1))" indicates that disturbing deep-level features can introduce excessive disruptions, increasing training complexity. Disturbing only the shallow-level features appears more robust than disturbing the deep-level features.
\ding{185} Under the ``With Gate" condition, ``Ours(FP(d1))" employs ResNet50 as the backbone. In comparison with the approach based on ``ResNet101", it is observed that with a limited number of labels, the performance of the ResNet50-based is slightly better than that of the ResNet101-based. As the number of labels increases, the performance of ResNet101-based  surpasses that of the ResNet50-based.

\begin{figure}[!ht]
\centering
\centerline{\includegraphics[width=0.85\linewidth]{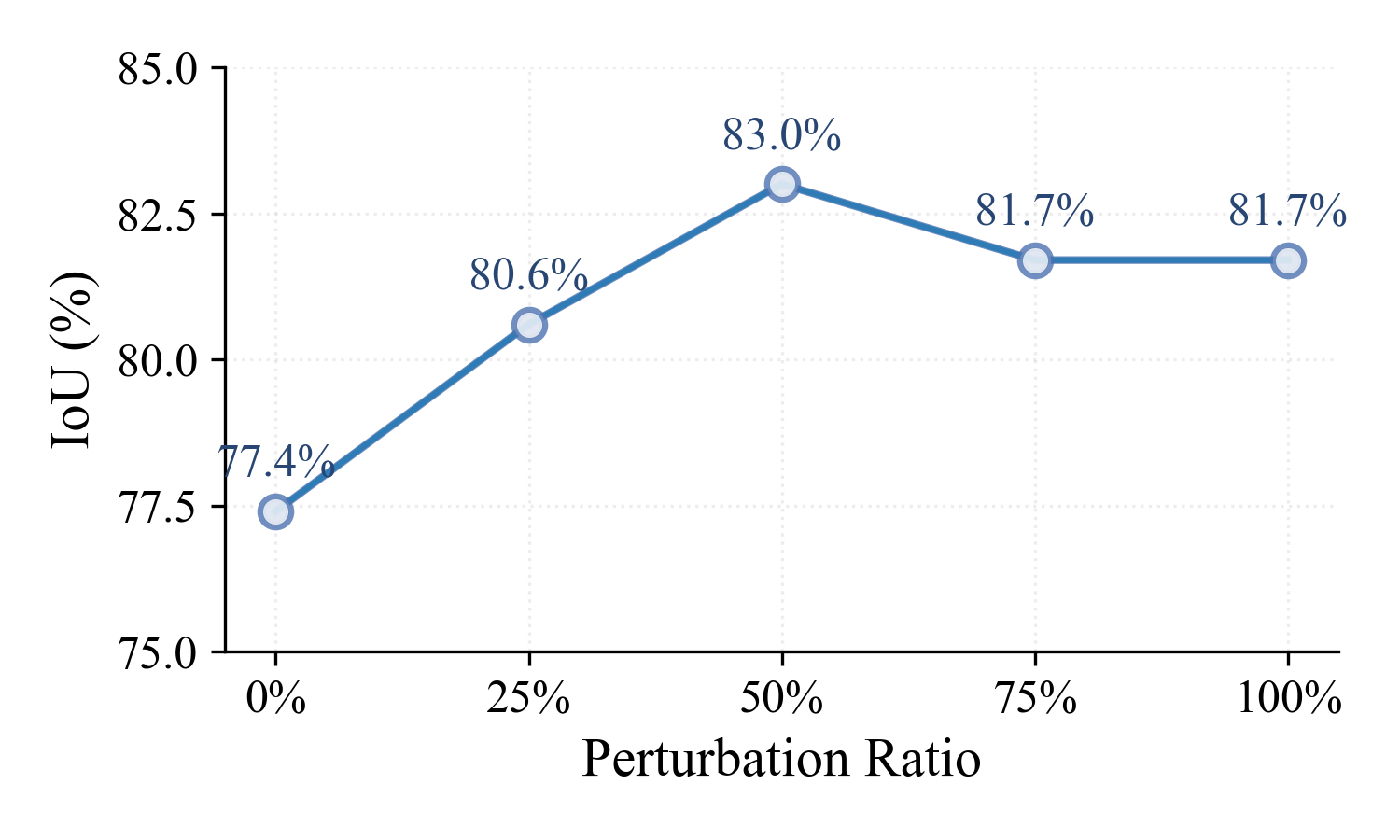}}
\vspace{-10px}
\caption{Comparison of different perturbation ratios at \(5\%\) labeled WHU-CD.}
\label{fig:visiual_gate}
\end{figure}

\textbf{Sensitivity of the gating mechanism.}
Fig.~\ref{fig:visiual_gate} shows the perturbation ratio experiments on WHU-CD with \(5\%\) labeled data.
A high ratio leads to unstable training as more samples are perturbed, while a low ratio may miss perturbation opportunities. The median ratio achieves a proper balance and attains the best performance.

\section{Conclusion}
In this work, we introduce the GTPC-SSCD, a novel semi-supervised change detection method based on gate-guided two-level perturbation consistency regularization. It maintains both image-level strong-to-weak consistency and feature-level perturbation consistency, facilitating efficient utilization of the abundant unlabeled data.
We also develop a hardness analysis-based gating mechanism to intelligently determine the necessity of implementing feature perturbations per sample. 
Extensive experiments on six benchmark change detection datasets validate the effectiveness and superiority of our proposed method.
In the future, we plan to explore network autonomy in perturbation strategy selection for diverse data scenarios to boost the performance and applicability.

\bibliographystyle{ieeetr}
\bibliography{icme2025}

\end{document}